\documentclass[11pt]{article}

\usepackage[utf8]{inputenc} % allow utf-8 input
\usepackage[T1]{fontenc}    % use 8-bit T1 fonts
\usepackage{hyperref}       % hyperlinks
\usepackage{url}            % simple URL typesetting
\usepackage{booktabs}       % professional-quality tables
\usepackage{amsfonts}       % blackboard math symbols
\usepackage{nicefrac}       % compact symbols for 1/2, etc.
\usepackage{microtype}      % microtypography
\usepackage[top=25truemm,bottom=25truemm,left=25truemm,right=25truemm]{geometry}

\usepackage{color}
\usepackage[dvipdfmx]{graphicx}
\usepackage{algorithm}
\usepackage{algorithmic}
\usepackage{indentfirst}
\usepackage{amsthm}
\usepackage{amsmath}

\title{{\Large Knowledge Discovery from Layered Neural Networks\\based on Non-negative Task Decomposition}}

\author{{\normalsize Chihiro Watanabe}\thanks{\textit{Email address:} watanabe.chihiro@lab.ntt.co.jp}{\normalsize ,\ \ Kaoru Hiramatsu,\ \ Kunio Kashino}\\
{\small\itshape NTT Communication Science Laboratories,}\\{\small\itshape 3-1, Morinosato Wakamiya, Atsugi-shi, Kanagawa Pref. 243-0198 Japan}}
\date{}

\begin{document}

\maketitle

\begin{abstract}
Interpretability has become an important issue in the machine learning field, along with the success of layered neural networks in various practical tasks. 
Since a trained layered neural network consists of a complex nonlinear relationship between large number of parameters, we failed to understand 
how they could achieve input-output mappings with a given data set. 
In this paper, we propose the non-negative task decomposition method, which applies non-negative matrix factorization to a trained layered neural network. 
This enables us to decompose the inference mechanism of a trained layered neural network into multiple principal tasks of input-output mapping, 
and reveal the roles of hidden units in terms of their contribution to each principal task.
\end{abstract}

%##################################################################################################################################################################################

\section{Introduction}

The interpretability of machine learning models or their trained result has become an important issue, along with the recent success of layered neural networks (or LNN). 
Their complex hierarchical network structures have made it possible to represent nonlinear complex relationships between input and output data, and greatly improve prediction accuracy 
with various practical data sets \cite{Krizhevsky2012,Tompson2014,Hinton2012,Sainath2013,Sutskever2014,Collobert2011}. 
Despite this powerful prediction ability, their black-box inference mechanism has limited their application area. 
For instance, in the area of automatic driving or medical care, 
a reasonable explanation must be provided as to how a trained LNN derived a prediction result. 

Recently, various approaches have been proposed to solve such problems of interpretability. 
For instance, a method has been proposed that first selects important features of data, and then approximates the trained LNN with an interpretable function such as a linear 
model \cite{Lundberg2017,Nagamine2017,Ribeiro2016}. 
There are also studies that analyze the effect of each unit output on inference results \cite{Raghu2017,Zahavy2016,Luo2016,Li2015}, 
or simplify the structure of trained networks by decomposing the units into clusters or communities by employing network analysis \cite{Watanabe2018,Watanabe2017b,Watanabe2017d}. 
Other studies have explored LNN training methods for constructing a trained network that is represented by an interpretable function \cite{Gonzalez2017,Foerster2017}. 

These studies have enabled us to obtain knowledge about various different aspects of an LNN, however, no method has been developed for simplifying the LNN structure 
by decomposing the hidden units into communities and simultaneously revealing the role of each community in inference. 

In this paper, we propose the non-negative task decomposition of LNNs, which enables us to obtain a simplified global structure of a trained LNN and 
knowledge about the function of each decomposed part, simultaneously. 
Unlike the methods described in previous studies \cite{Watanabe2018,Watanabe2017b,Watanabe2017d} for analyzing trained LNNs by detecting their layer-wise community structure 
via post-processing, 
our proposed method can reveal both a set of principal input-output mapping tasks and the community structure of hidden units across layers at the same time. 

To achieve such task decomposition, we first determine the role of each hidden unit as a vector that represents the effect of each input dimension on the hidden unit and 
the effect of the hidden unit on each output dimension. Then, we apply non-negative matrix factorization to a matrix consisting of such vectors for all the hidden units, and 
obtain information about both the principal tasks in a trained LNN and the classification result of hidden units. 
A related work has been proposed for representing the role of each unit as a vector \cite{Raghu2017}, however, our proposed method differs from the previous study in that 
it can break down both the function and structure of an LNN into multiple components, providing the inference organization of a trained LNN. 

We show experimentally that our proposed method can reveal the inference mechanism of an LNN. 
First, we apply our proposed method to an LNN trained with a data set with a ground truth community structure, and show that the hidden units are appropriately decomposed 
into communities across layers. Then, we analyze LNNs trained with a sequential data set and an image data set by employing our proposed method, and discuss the results. 

%##################################################################################################################################################################################

\section{Training Layered Neural Networks based on Back Propagation}

We first train an LNN with an arbitrary data set by the same method as that used in previous studies \cite{Watanabe2018,Watanabe2017b,Watanabe2017d}. 
Let $(x, y)$, $x\in \mathbb{R}^M$, $y\in \mathbb{R}^N$ be a data set consisting of input data $x$ and output data $y$, and let $q(x,y)$ be 
a probability density function on $\mathbb{R}^M\times \mathbb{R}^N$ for training and test data sets. 
When using an LNN, we assume that the relationship between input data and output data is represented by a function $f(x,w)$ 
from $x\in \mathbb{R}^M,\ w\in \mathbb{R}^L$ to $\mathbb{R}^N$ that predicts output data $y$ from input data $x$ and a parameter $w=\{\omega^d_{ij}, \theta^d_i\}$, 
where $\omega^d_{ij}$ represents the connection weight between the $i$-th unit in a depth $d$ layer and the $j$-th unit in a depth $d+1$ layer, and 
$\theta^d_i$ represents the bias of the $i$-th unit in the depth $d$ layer. 
Here, the depth $1$ and $D$ layers correspond to the input and output layers, respectively. 
The function $f_j(x,w)$ of an LNN for the $j$-th unit in the output layer is given by $f_j(x,w) = \sigma(\sum_i \omega^{D-1}_{ij} o^{D-1}_i+\theta^{D-1}_j)$, 
where
\begin{eqnarray*}
  o^{D-1}_j = \sigma(\sum_i \omega^{D-2}_{ij} o^{D-2}_i+\theta^{D-2}_j),\ \ \ \ \ \ \cdots, \ \ \ \ \ \ o^2_j = \sigma(\sum_i \omega^1_{ij} x_i +\theta^1_j),
\end{eqnarray*}
and $\sigma (x)=1/(1+\exp (-x))$. 
We train an LNN to determine the parameter $w$ of the above function with a training data set $\{(X_n,Y_n)\}_{n=1}^{n_1}$ with sample size $n_1$. 

The training error $E(w)$ and the generalization error $G(w)$ of an LNN, respectively, are given by 
\begin{eqnarray*}
  E(w) = \frac{1}{n_1} \sum_{n=1}^{n_1} \|Y_n-f(X_n,w)\|^2,\ \ \ \ \ \ \ \ \ 
  G(w) = \int \|y-f(x,w)\|^2 q(x,y)dxdy,
\end{eqnarray*}
where $\|\cdot\|$ is the Euclidean norm of $\mathbb{R}^N$. 
Since the generalization error cannot be calculated with a finite size data set, we approximate it by $G(w)\approx \frac{1}{m_1} \sum_{m=1}^{m_1} \|{Y_m}'-f({X_m}',w)\|^2$, 
where $\{({X_m}', {Y_m}')\}_{m=1}^{m_1}$ is a test data set that is independent of the training data set. 

In this paper, to avoid overfitting to a training data set, we adopt the LASSO method (\cite{Ishikawa1990,Tibshirani1996}) to automatically delete small connection weights, 
where the objective function to be minimized is given by $H(w) = \frac{n_1}{2}\ E(w)+\lambda \sum_{d,i,j} |\omega^d_{ij}|$, 
where $\lambda$ is a hyperparameter. 
The value of the above function $H(w)$ is minimized with the stochastic steepest descent method, where the parameters are iteratively updated with the following equations 
(\cite{Werbos1974,Rumelhart1986}). 

For the $D$-th layer, 
\begin{eqnarray*}
  &&\Delta \omega^{D-1}_{ij} = -\eta (\delta^{D}_j o^{D-1}_i+\lambda \ \mathrm{sgn}(\omega^{D-1}_{ij})),\ \ \ \ \ \ \ \ \ 
  \Delta \theta^D_j = -\eta \delta^{D}_j,
\end{eqnarray*}
where $\delta^{D}_j = (o^{D}_j-y_j)\ (o^{D}_j\ (1-o^{D}_j) +\epsilon_1)$. 

For $d=D-1,\ D-2,\cdots, 2$, 
\begin{eqnarray*}
  &&\Delta \omega^{d-1}_{ij} = -\eta (\delta^d_j o^{d-1}_i+\lambda \ \mathrm{sgn}(\omega^{d-1}_{ij})),\ \ \ \ \ \ \ \ \ 
  \Delta \theta^d_j = -\eta \delta^d_j,
\end{eqnarray*}
where $\delta^d_j = \sum_{k=1}^{l_{d+1}} \delta^{d+1}_k \omega^d_{jk}\ (o^d_j\ (1-o^d_j) +\epsilon_1)$. 

Here, $y_j$ is the $j$-th dimension value of the output data of the randomly chosen $n$-th sample $(X_n, Y_n)$, 
$\epsilon_1$ is a hyperparameter for the LNN convergence, and 
$\eta$ for training time $t$ is defined such that $\eta(t)\propto 1/t$. 
In this paper, we defined $\eta =0.7\times a_1 n_1/(a_1 n_1 +5t)$, where $a_1$ is the mean iteration number for LNN training per dataset. 

%##################################################################################################################################################################################

\section{Knowledge Discovery from Layered Neural Networks based on Non-negative Task Decomposition}

\subsection{Extracting Feature Vectors of Hidden Layer Units Based on Their Relationship with Input and Output Dimensions}
\label{sec:feature}

To decompose the function of a trained LNN, we first define a non-negative matrix $V=\{v_{k,l}\}$, 
whose $k$-th row consists of a feature vector $v_k$ of the $k$-th unit in a hidden layer. 
Here, we define the feature vector $v_k$ by using a method described in a previous study \cite{Watanabe2018arxiv} for determining quantitatively the role of each community 
(or cluster) of units as regards each unit in an LNN. 
In the previous study, the role of community $c$ is given by a pair of feature vectors $v^{\mathrm{in}}_{c}=\{v^{\mathrm{in}}_{ic}\}$ and 
$v^{\mathrm{out}}_{c}=\{v^{\mathrm{out}}_{cj}\}$, which represent the effect of the $i$-th input dimension on the community $c$ and 
the effect of the community $c$ on the $j$-th output dimension, respectively. 
With our proposed method, we assume that each LNN community consists of a single hidden unit, and we determine a pair of feature vectors for that unit by 
employing the method used in \cite{Watanabe2018arxiv}. 

Specifically, the effect of the $i$-th input dimension on the $k$-th hidden unit is computed as the square root error of the output in the $k$-th hidden unit, 
when the $i$-th input dimension is replaced with the mean value for the training data (in other words, when the LNN cannot use the value of the $i$-th 
input dimension). 
This definition is given by the following equations. 

\textbf{Effect of $i$-th input dimension on $k$-th hidden unit: }
Let $o^{(n)}_k$ be the output of the $k$-th hidden unit for the $n$-th input data sample $X^{(n)}$, 
and let $z^{(n)}_k$ be the output of the $k$-th hidden unit for an input data sample $X'^{(n)}$ that is generated based on the following definition: 
\begin{eqnarray*}
  X'^{(n)}_i \equiv \frac{1}{n_1} \sum_n X^{(n)}_i. \\
  \mathrm{For\ } l\neq i,\ X'^{(n)}_l \equiv X^{(n)}_l. 
\end{eqnarray*}
We define the effect of the $i$-th input dimension on the $k$-th hidden unit as 
\begin{eqnarray*}
v^{\mathrm{in}}_{ik}=\sqrt{\frac{1}{n_1} \sum_n \Bigl( o^{(n)}_k -z^{(n)}_k \Bigr)^2}. 
\end{eqnarray*}

Similarly, the effect of the $k$-th hidden unit on the $j$-th output dimension is computed as the square root error of the value of the $j$-th output dimension 
when the output in the $k$-th hidden unit is replaced with the mean value for the training data (in other words, when the LNN cannot use the value of the 
$k$-th hidden unit). 
This definition is given by the following equations. 

\textbf{Effect of $k$-th hidden unit on $j$-th output dimension}: 
Let $y^{(n)}_j$ be the output of the $j$-th unit in the output layer for the $n$-th input data sample $X^{(n)}$, 
and let $z^{(n)}_j$ be the output of the $j$-th unit in the output layer when changing the output values in the $k$-th hidden unit, according to the following procedure: 
We change the output value of the $k$-th hidden unit for the $n$-th input data sample from $o^{(n)}_k$ to $o'^{(n)}_k$. 
Here, $o'^{(n)}_k$ is given by 
\begin{eqnarray*}
  o'^{(n)}_{k} \equiv \frac{1}{n_1} \sum_n o^{(n)}_{k}. 
\end{eqnarray*}
We define the effect of the $k$-th hidden unit on the $j$-th output dimension as 
\begin{eqnarray*}
v^{\mathrm{out}}_{kj}=\sqrt{\frac{1}{n_1} \sum_n \Bigl( y^{(n)}_j -z^{(n)}_j \Bigr)^2}. 
\end{eqnarray*}

Once we have obtained the feature vectors $v^{\mathrm{in}}_{k}$ and $v^{\mathrm{out}}_{k}$ for the $k$-th hidden unit, we define a single feature vector $v_k$ 
by combining these two kinds of feature vectors. 
In this paper, before combining the feature vectors, we normalize them so that the minimum and maximum values are the same for vectors $v^{\mathrm{in}}_{k}$ and 
$v^{\mathrm{out}}_{k}$ when changing the hidden unit index $k$. 
Specifically, we define the minimum and maximum values of $v^{\mathrm{in}}_{k}$ and $v^{\mathrm{out}}_{k}$ for all the hidden units as $0$ and $1$, respectively. 
This normalization is undertaken so that the effect of an input dimension on a hidden unit and the effect of a hidden unit on an output dimension are treated equally. 
We define a feature vector $v_k=\{v_{k,l}\}$ constituting a non-negative matrix $V$ by the following equations. 
\begin{eqnarray*}
  \mathrm{For\ } 1\leq l\leq i_0,\ v_{k,l} &\equiv& v^{\mathrm{in}}_{k,l}, \\
  \mathrm{For\ } i_0 +1\leq l\leq i_0 +j_0 ,\ v_{k,l} &\equiv& v^{\mathrm{out}}_{k,l-i_0}, 
\end{eqnarray*}
where $i_0$ and $j_0$, respectively, represent the dimensions of the input and output data. 

%========================================================================================================================================================================

\subsection{Non-negative Task Decomposition of Layered Neural Networks}

\begin{figure}[t]
    \begin{tabular}{cc}
      \begin{minipage}[b]{0.45\hsize}
        \centering
        \includegraphics[height=45mm]{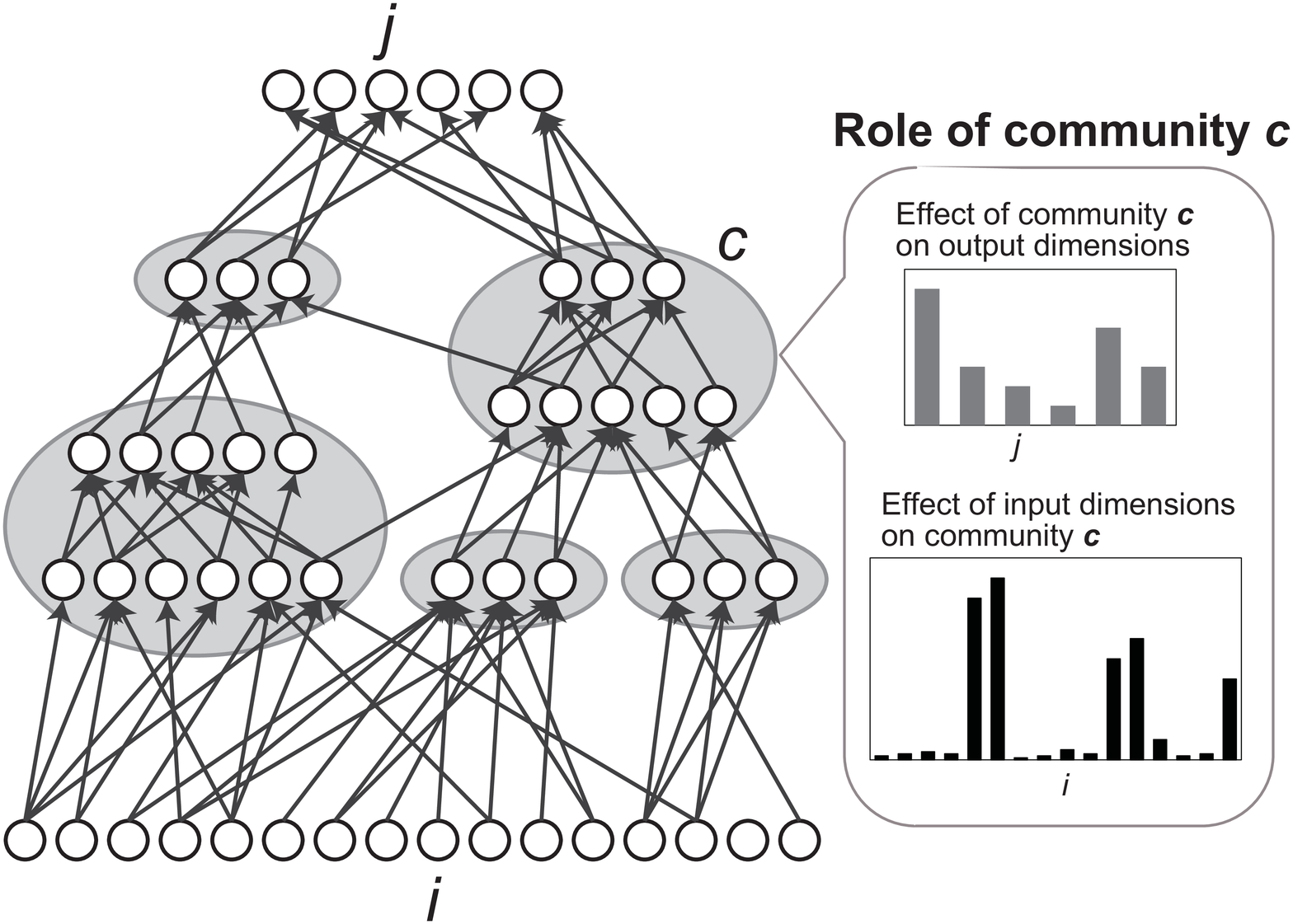}
      \end{minipage} &
      %\hspace{5mm}
      \begin{minipage}[b]{0.45\hsize}
        \centering
        \includegraphics[height=40mm]{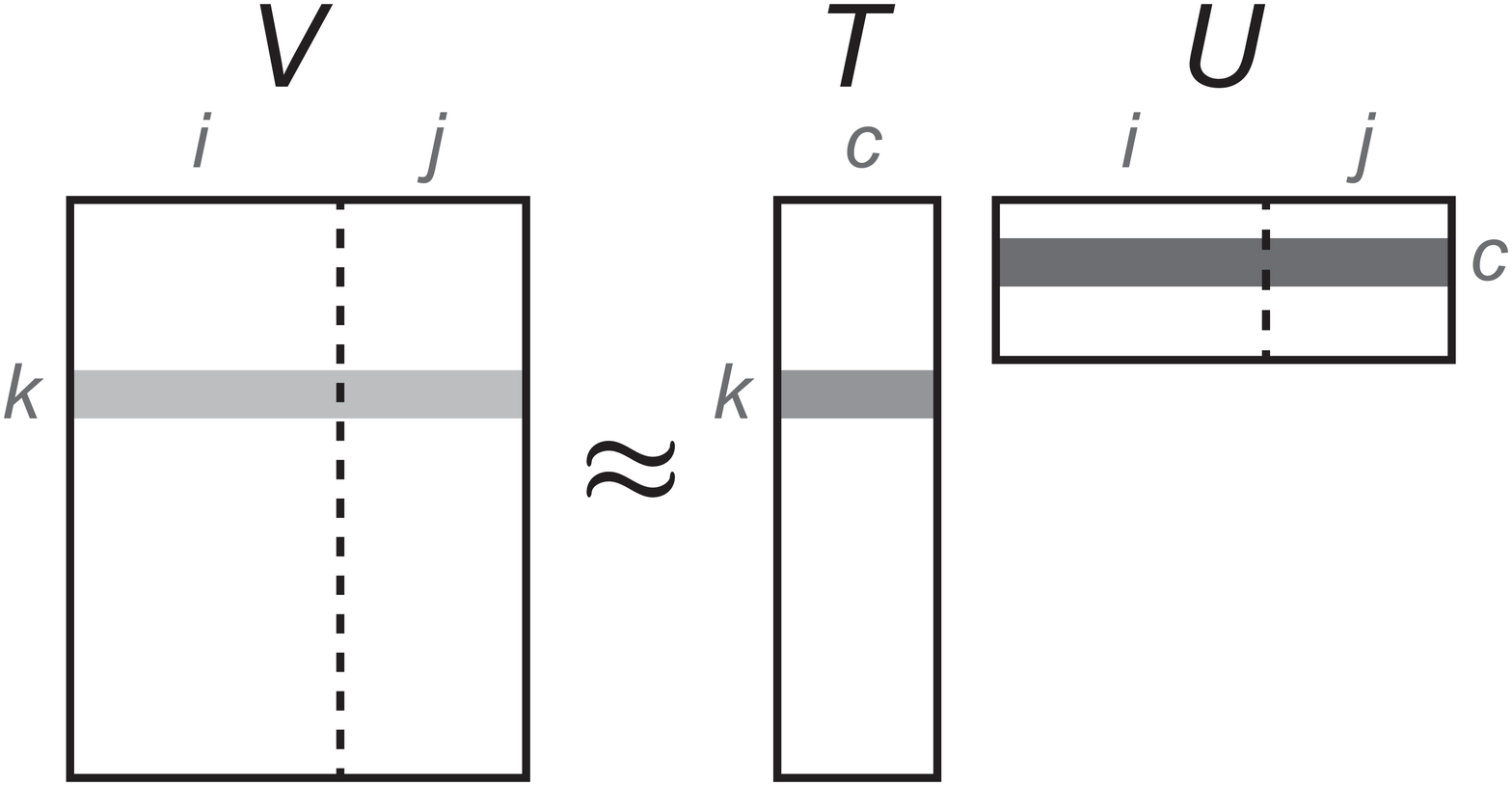}\vspace{5mm}
      \end{minipage}
    \end{tabular}
  \vspace{-5mm}
  \caption{Non-negative task decomposition of an LNN. 
  \textbf{Left}: Hidden units are decomosed into multiple communities, according to the their roles in inference. 
  Our proposed method reveals the main role of each community in terms of the strength of the relationship with each input and output dimension. 
  \textbf{Right}: Non-negative task decomposition is achieved by approximating the non-negative matrix $V$ with the product of low-dimensional non-negative matrices $T$ and $U$. 
  Here, the $k$-th row of the matrix $V$ represents the role of the $k$-th hidden unit in terms of the relationship with each input dimension $i$ and each output dimension $j$. 
  The $k$-th row of the matrix $T$ corresponds to the clustering result of the $k$-th hidden unit, 
  which is represented as a weight for each community $c$. 
  The $c$-th row of the matrix $U$ corresponds to the $c$-th decomposed task of the LNN. }
  \label{fig:nnmf}
\end{figure}

Here, we describe a method for decomposing the function of whole hidden layers of a trained LNN into multiple main tasks.
As shown in Figure \ref{fig:nnmf}, this LNN task decomposition is achieved by approximating a non-negative matrix $V$ consisting of the feature vectors of hidden units 
described in section \ref{sec:feature} with the product of low-dimensional non-negative matrices $T=\{t_{k,c}\}\in {\mathbb{R}^+}^{k_0 \times c_0}$ and 
$U=\{u_{c,l}\}\in {\mathbb{R}^+}^{c_0 \times (i_0 +j_0)}$, where $k_0$ is the number of hidden units and $c_0$ is a hyperparameter. 
By such approximation, we represent the mapping from input dimension values to output dimension values by a hidden unit as the weighted linear sum of $c_0$ representative vectors. 
The $c$-th row of the matrix $U$ corresponds to the $c$-th representative vector, which represents a main task or input-output mapping that is performed by the hidden units in 
a trained LNN. 
On the other hand, the $k$-th row of the matrix $T$ corresponds to the weights of the representative vectors that constitute the task of the $k$-th hidden unit. 
The values of the matrix $T$ provide us with information regarding the similarity between tasks performed by different hidden units. 

The above non-negative approximation $V\approx TU$ is achieved by non-negative matrix factorization performed with Algorithm \ref{alg_NNMF} \cite{Lee2000}. 
By iteratively updating the values of matrices $T$ and $U$, we can obtain a local optimal solution for the approximation. 

%o o o o o o o o o o o o o o o o o o o o o o o o o o o o o o o o o o o o o o o o
\begin{algorithm}	
\caption{Non-negative Task Decomposition of Layered Neural Networks}
\label{alg_NNMF}
\begin{algorithmic}[1]
\STATE Let $a_0$ be the number of iterations of the algorithm, and let $V$ be a non-negative matrix consisting of feature vectors of hidden units. 
\STATE Initialization of matrices $T$ and $U$: $t_{k,c}\overset{\text\small\rm{i.i.d.}}{\sim}\mathcal{N}(\mu_1,\sigma_1)$, 
and $u_{c,l}\overset{\text\small\rm{i.i.d.}}{\sim}\mathcal{N}(\mu_2,\sigma_2)$. In this paper, we set the values at $\mu_1=\sigma_1=\mu_2=\sigma_2=0.5$. 
\FOR{$a=1$ to $a_0$}
  \STATE $t_{k,c}\gets t_{k,c}\times ((VU^{T})_{k,c}/(TUU^{T})_{k,c})$.
  \STATE $t_{k,c}\gets \max (t_{k,c}, 0)$.
  \STATE $u_{c,l}\gets u_{c,l}\times ((T^{T}V)_{c,l}/(T^{T}TU)_{c,l})$.
  \STATE $u_{c,l}\gets \max (u_{c,l}, 0)$.
\ENDFOR
\end{algorithmic}
\end{algorithm}
%o o o o o o o o o o o o o o o o o o o o o o o o o o o o o o o o o o o o o o o o
With the above non-negative matrix factorization, we can gain knowledge about the combination of tasks that mainly constitutes the inference mechanism of the entire trained LNN. 

%o o o o o o o o o o o o o o o o o o o o o o o o o o o o o o o o o o o o o o o o
%o o o o o o o o o o o o o o o o o o o o o o o o o o o o o o o o o o o o o o o o

%##################################################################################################################################################################################

\section{Experiments}
\label{sec:experiment}

We show the effectiveness of our proposed method experimentally by using three types of data sets, which are the same as those used in \cite{Watanabe2018arxiv}. 

\subsection{Preliminary Experiment Using Synthetic Data Set}

First, we show that our proposed method can successfully decompose the tasks of a trained LNN by using a synthetic data set with a ground truth community structure. 
We define the ground truth structure as constituting three independent LNNs, each of which has two hidden layers that each include $15$ units. 
The parameters $\hat{w}=\{\hat{\omega}^d_{ij}, \hat{\theta}^d_i\}$ for the ground truth LNN are generated by 
$\hat{\omega}^d_{ij}\overset{\text\small\rm{i.i.d.}}{\sim}\mathcal{N}(0,1)$, and $\hat{\theta}^d_i\overset{\text\small\rm{i.i.d.}}{\sim}\mathcal{N}(0,0.5)$. 
Here, connection weights with absolute values of one or less were deleted. 

We train an LNN with the data set $\{(X_n,Y_n)\}$, which was generated by $X_n\overset{\text\small\rm{i.i.d.}}{\sim}\mathcal{N}(0,3)$, and 
$Y_n=f(X_n,\hat{w})+\epsilon_2$, where $\epsilon_2 \overset{\text\small\rm{i.i.d.}}{\sim}\mathcal{N}(0,0.05)$. 
Then, we apply our proposed non-negative task decomposition to the trained LNN. 

\begin{figure}[t]
  \centering
  \includegraphics[width=110mm]{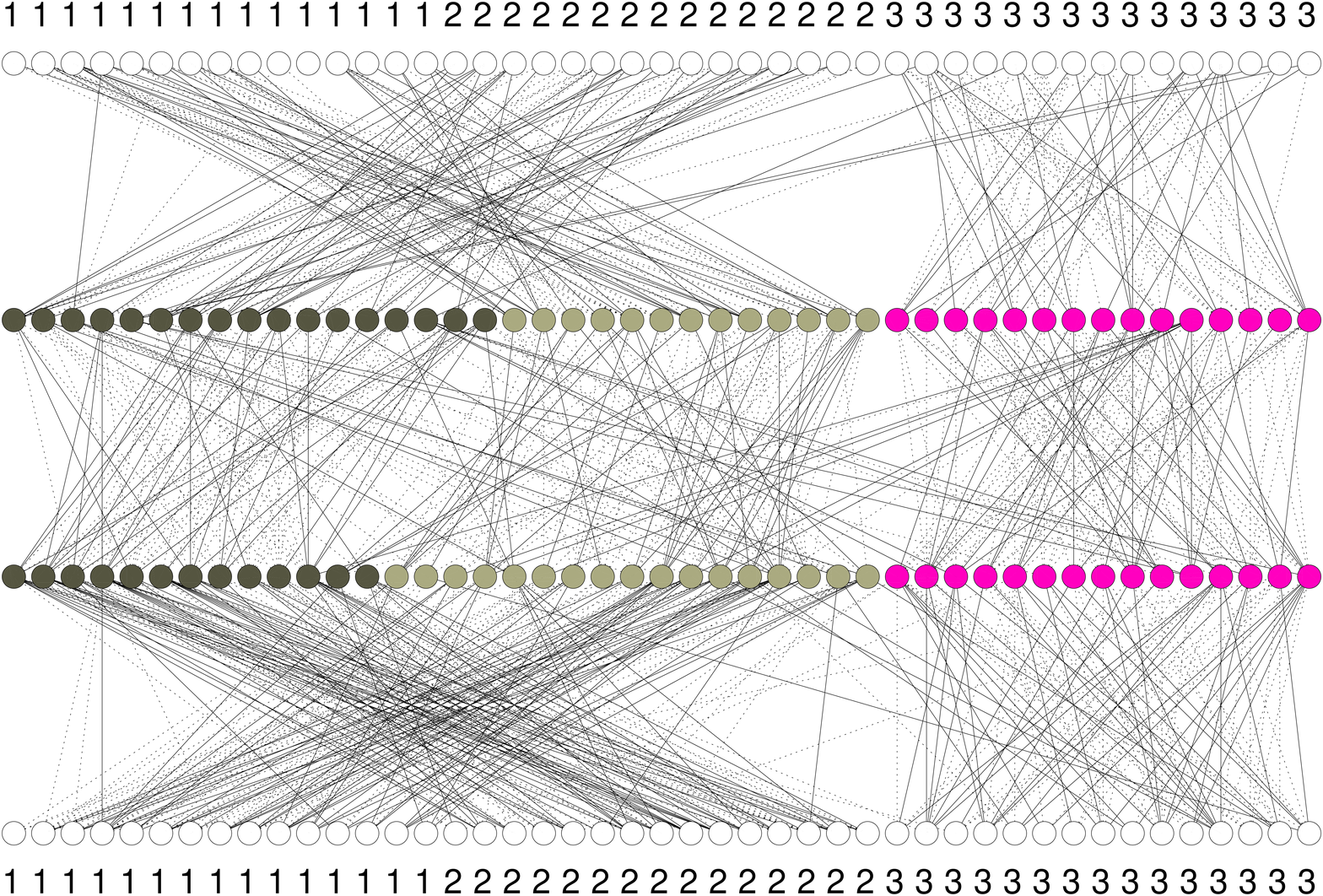}
  \caption{An LNN trained with a synthetic data set. 
        The numbers above the output layer and below the input layer show the community indices of the ground truth structure. 
        The solid and dotted lines, respectively, represent positive and negative connections (best viewed in color). }\vspace{10mm}
  \label{fig:syn_lnn}
  \includegraphics[width=105mm]{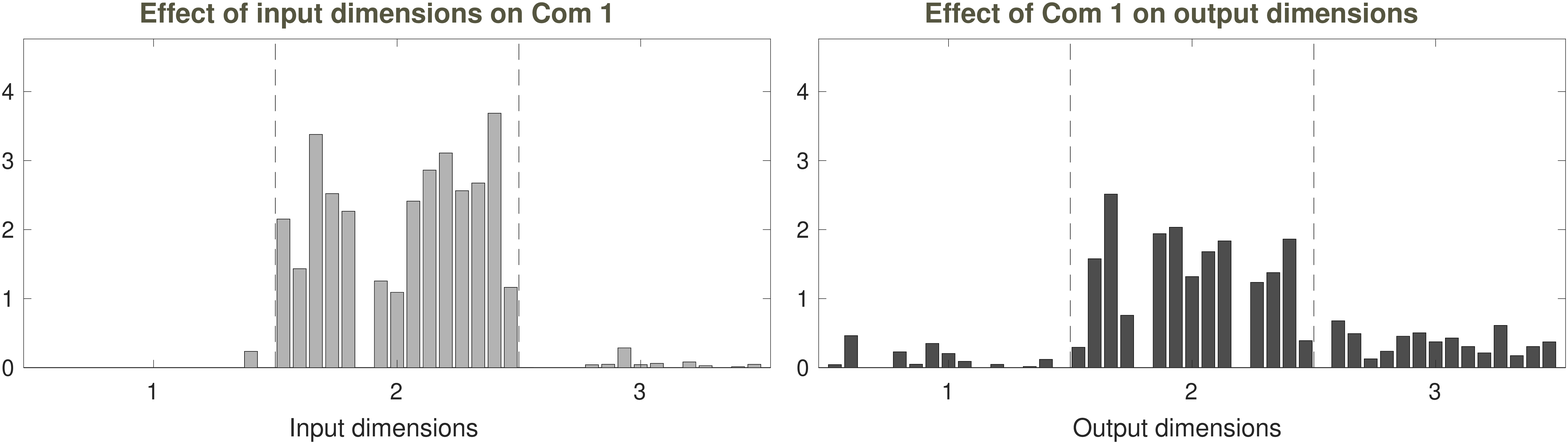}\\
  \includegraphics[width=105mm]{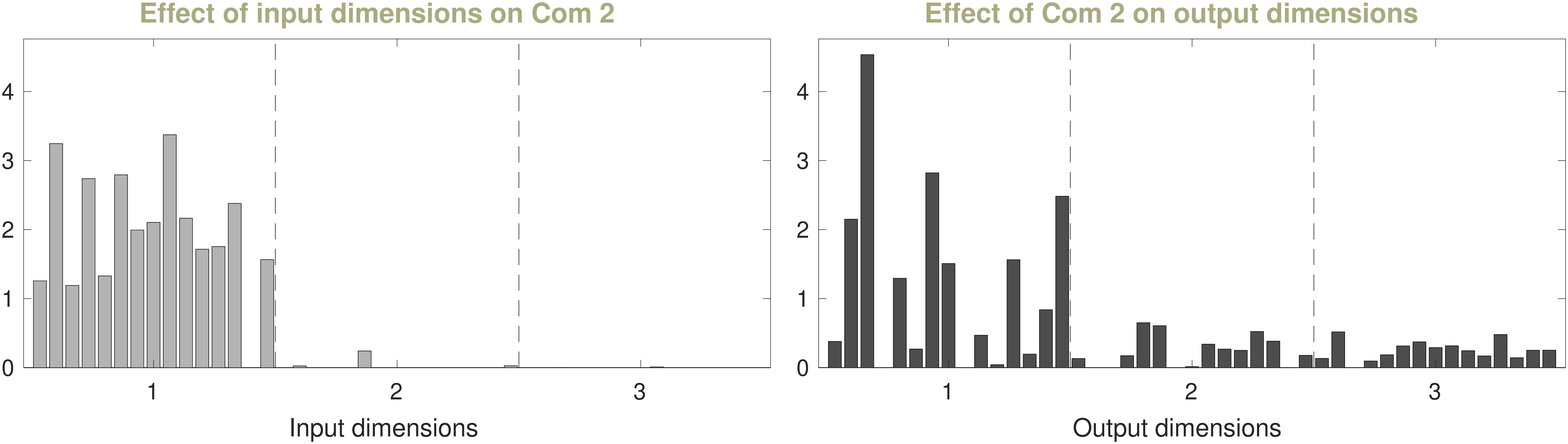}\\
  \includegraphics[width=105mm]{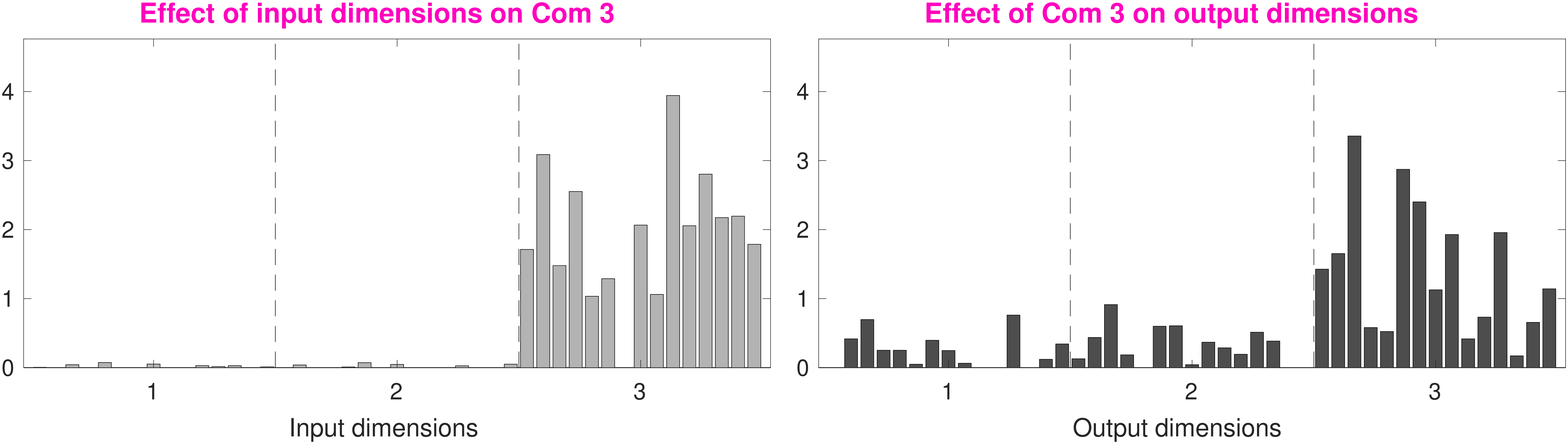}
  \caption{Decomposed tasks of an LNN trained with a synthetic data set. 
        The font colors of the figure titles correspond to the unit colors in Figure \ref{fig:syn_lnn} (best viewed in color). }
  \label{fig:syn_coms}
\end{figure}

Figures \ref{fig:syn_lnn} and \ref{fig:syn_coms}, respectively, show the trained LNN, and the decomposed tasks of the LNN. 
These figures show that our proposed method could appropriately decompose the trained LNN into three independent tasks, from the fact that each extracted community 
mainly performed mappings from input dimensions to output dimensions in one of the three independent networks. 
Here, note that unlike previous studies \cite{Watanabe2018,Watanabe2017b,Watanabe2017d} that decompose units layer-wise, 
our proposed method made it possible to classify the hidden units with similar functions into the same community, even if they were in different layers. 

\subsection{Experiment Using Sequential Data Set of Consumer Price Index}

Next, we trained an LNN with a sequential data set of a consumer price index \cite{estat} to predict the consumer price indices of taro, radish and carrot for a month 
from $36$ months' input data, and applied our proposed method to the trained LNN. 

\begin{figure}[t]
  \centering
  \includegraphics[width=110mm]{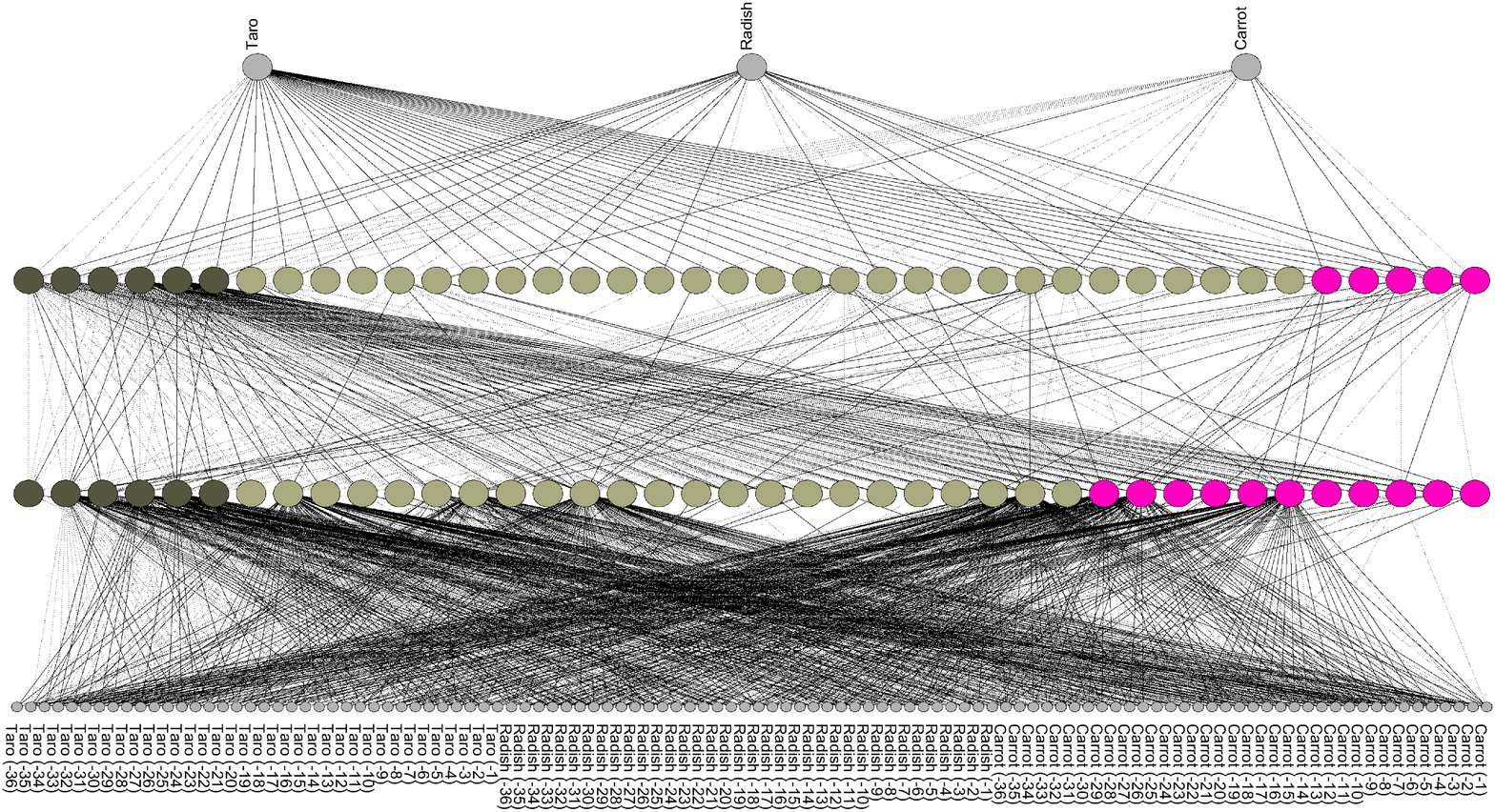}
  \caption{An LNN trained with a food consumer price index data set (best viewed in color). }\vspace{3mm}
  \label{fig:food_lnn}
  \fbox{\includegraphics[width=105mm]{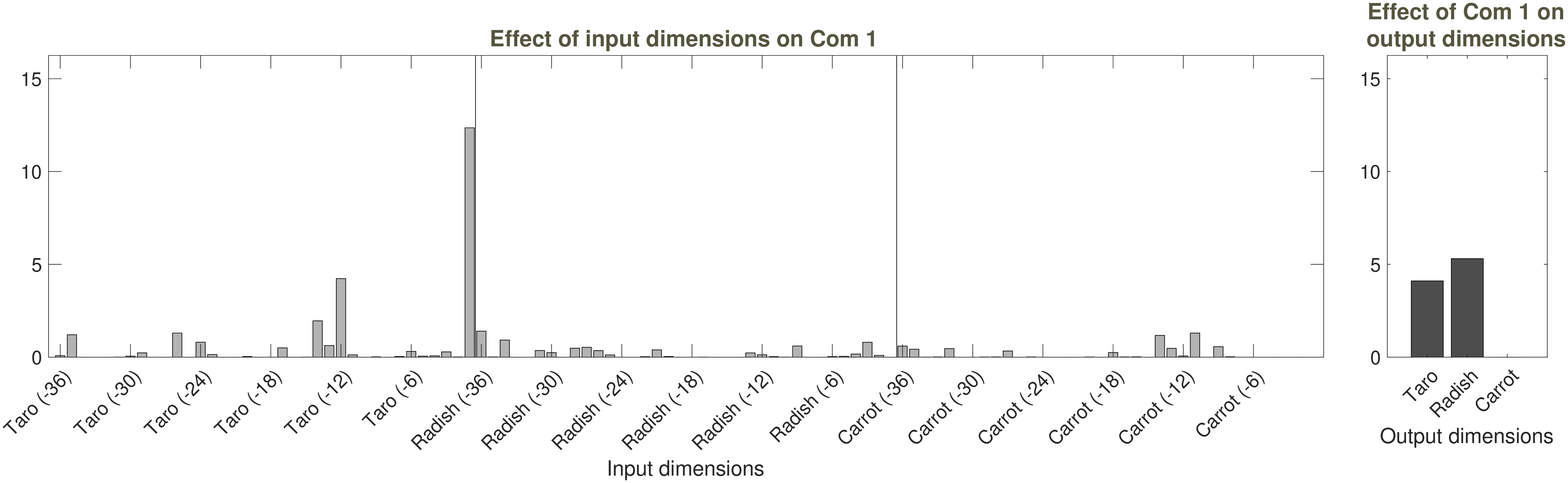}}\\
  \fbox{\includegraphics[width=105mm]{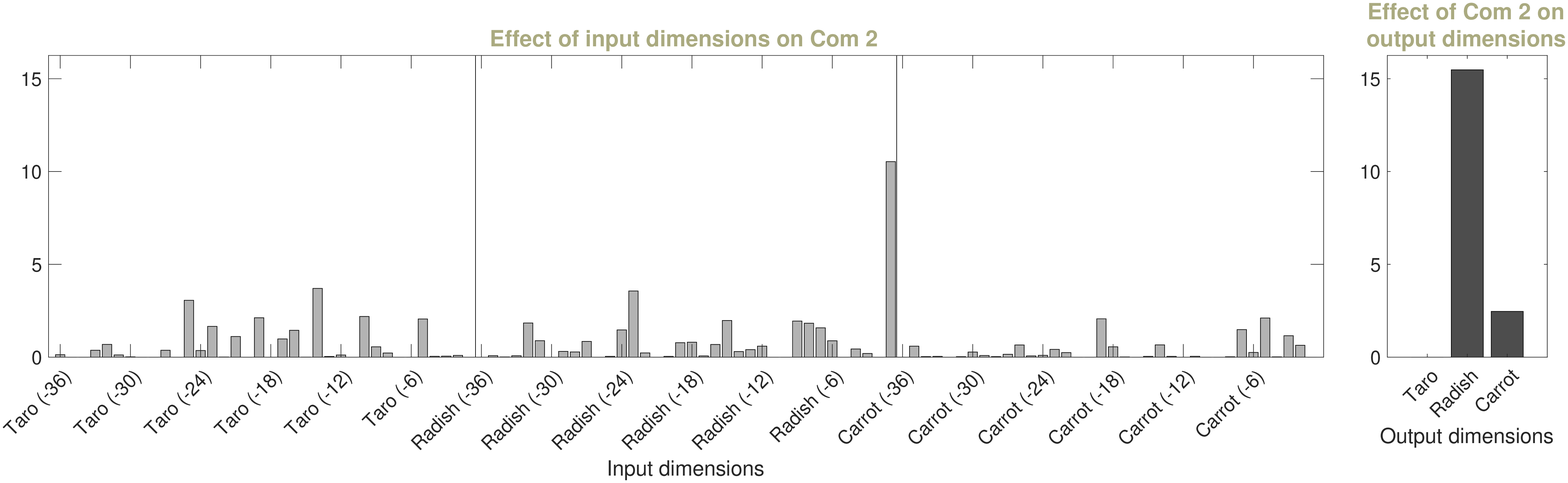}}\\
  \fbox{\includegraphics[width=105mm]{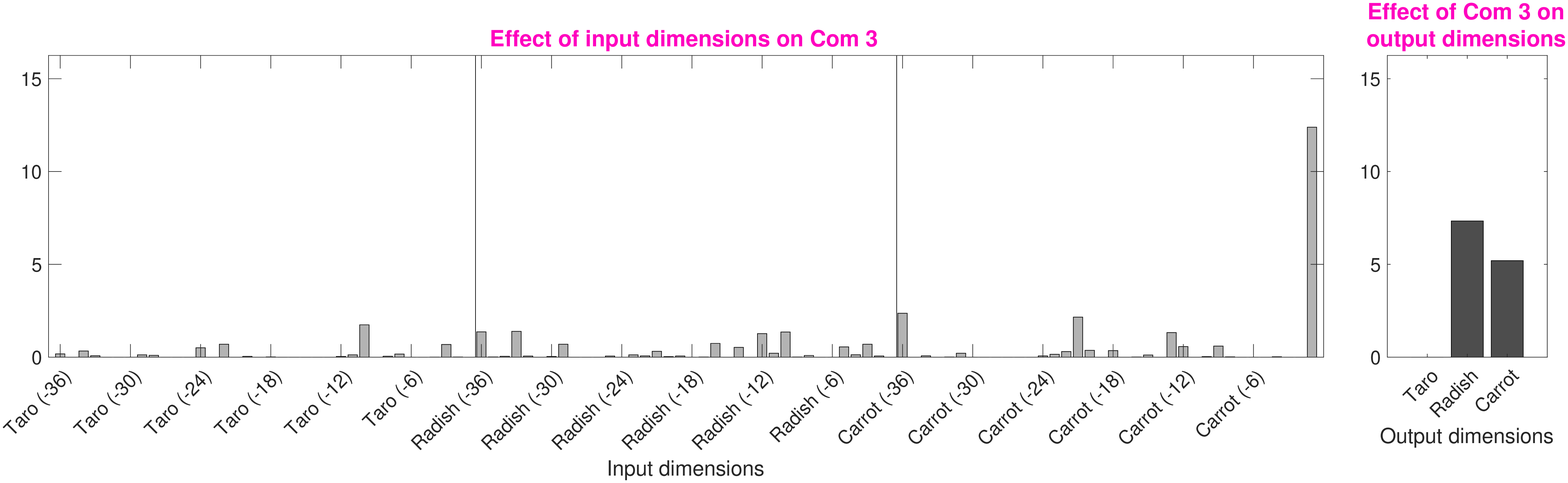}}
  \caption{Decomposed tasks of an LNN trained with a food consumer price index data set. 
  The font colors of the figure titles correspond to the unit colors in Figure \ref{fig:food_lnn} (best viewed in color). }
  \label{fig:food_coms}
\end{figure}

Figures \ref{fig:food_lnn} and \ref{fig:food_coms}, respectively, show the trained LNN, and the decomposed tasks of the LNN. 
From these figures, we can gain knowledge about the role of each community as follows. 

- Com $1$ predicts the consumer price indices for taro and radish mainly from information about taro from one month earlier and a year earlier. 

- Com $2$ predicts the consumer price index about radish mainly from information about radish from one and $23$ months earlier, and information about taro from $14$ 
and $25$ months earlier. 

- Com $3$ predicts the consumer price indices of radish and carrot mainly from information about carrot from one month earlier, $21$ months earlier, and three years earlier. 

%o o o o o o o o o o o o o o o o o o o o o o o o o o o o o o o o o o o o o o o o
%o o o o o o o o o o o o o o o o o o o o o o o o o o o o o o o o o o o o o o o o

\subsection{Experiment Using Diagram Image Data Set}

\begin{figure}[t]
  \centering
  %\hspace{20mm}
  \includegraphics[width=110mm]{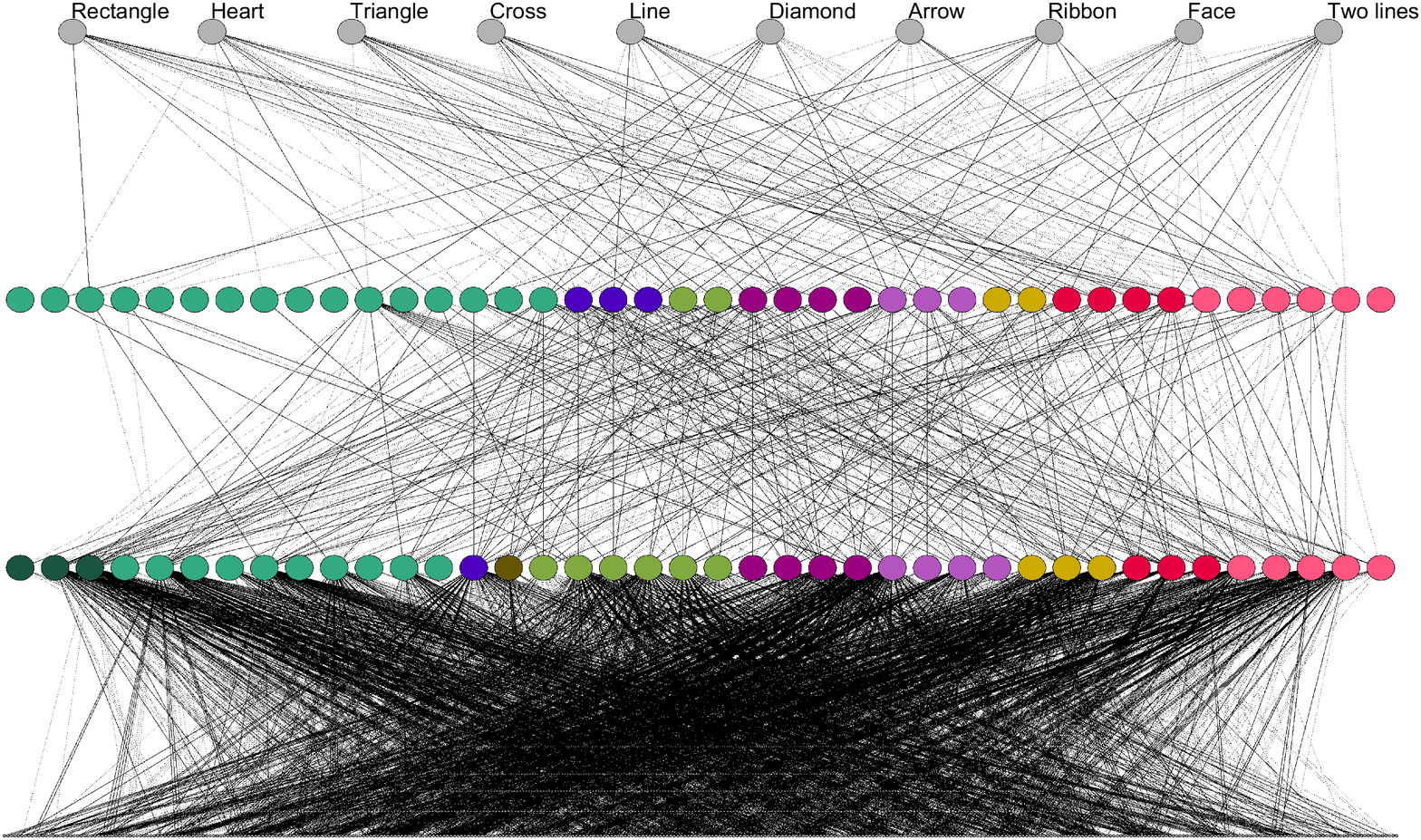}\vspace{-2mm}
  \caption{An LNN trained with a diagram image data set (best viewed in color). }\vspace{3mm}
  \label{fig:diagram_lnn}
  %\hspace{10mm}
  \fbox{\includegraphics[height=33mm]{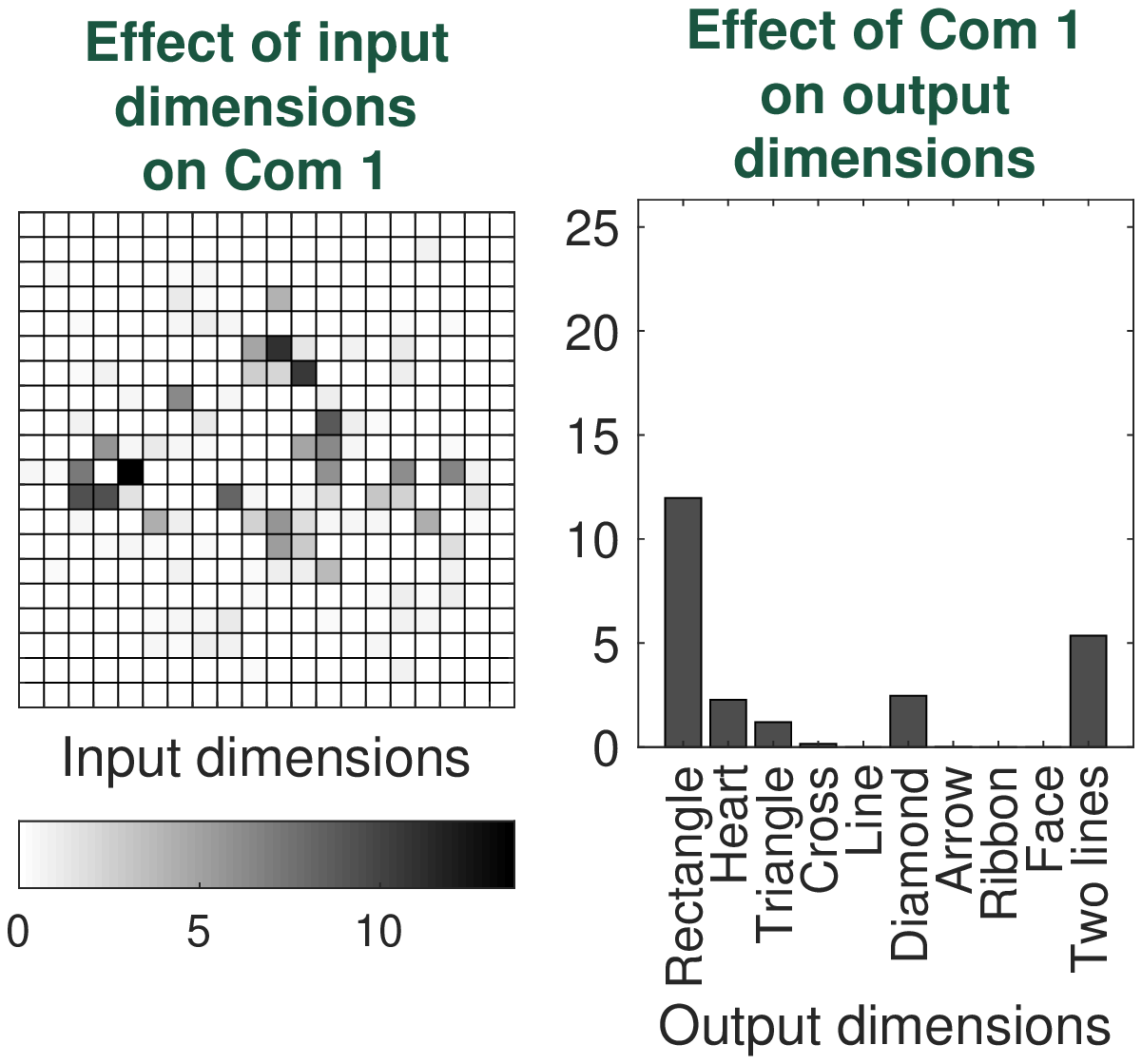}}
  \fbox{\includegraphics[height=33mm]{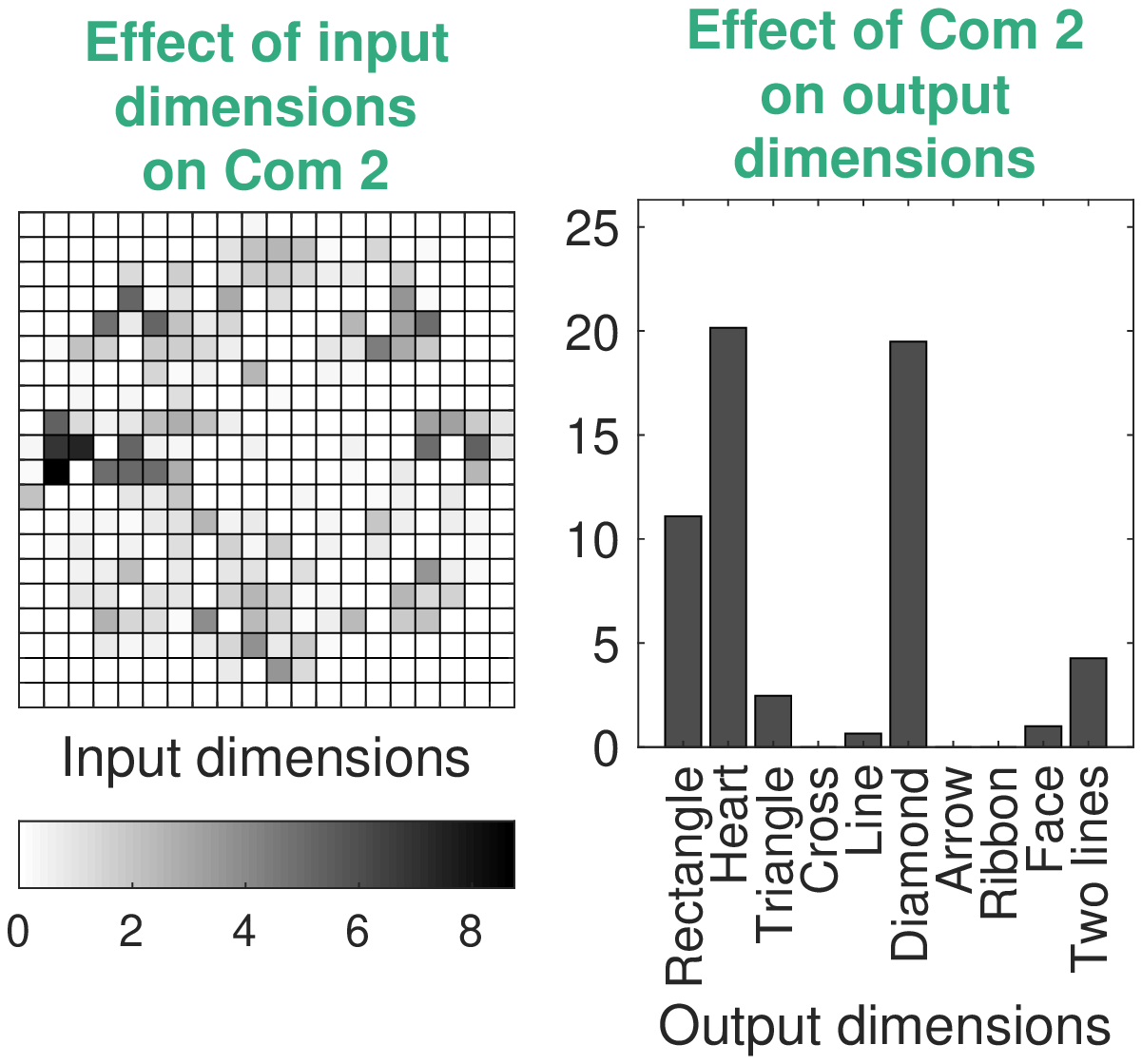}}
  \fbox{\includegraphics[height=33mm]{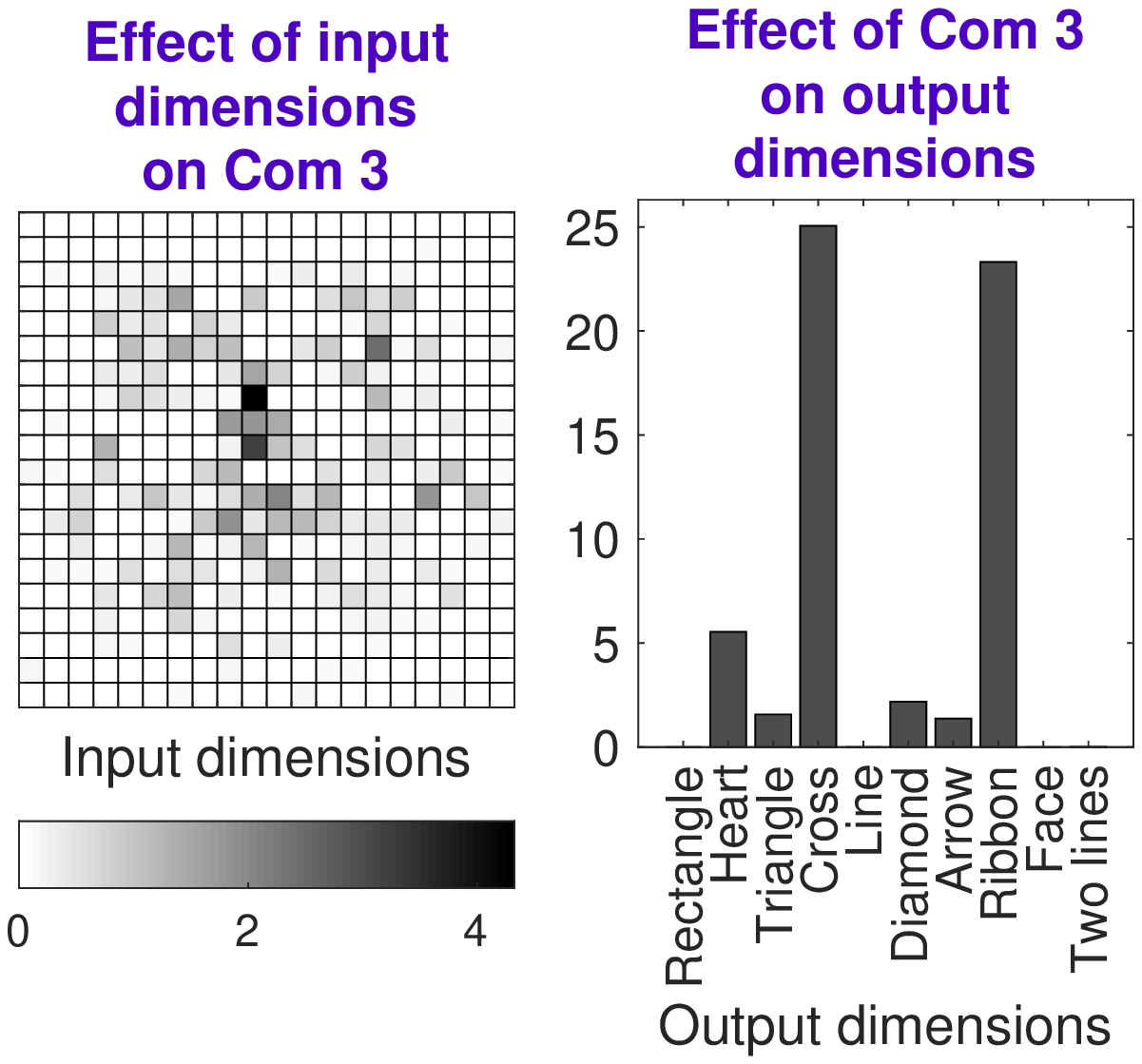}}\\
  %\hspace{10mm}
  \fbox{\includegraphics[height=33mm]{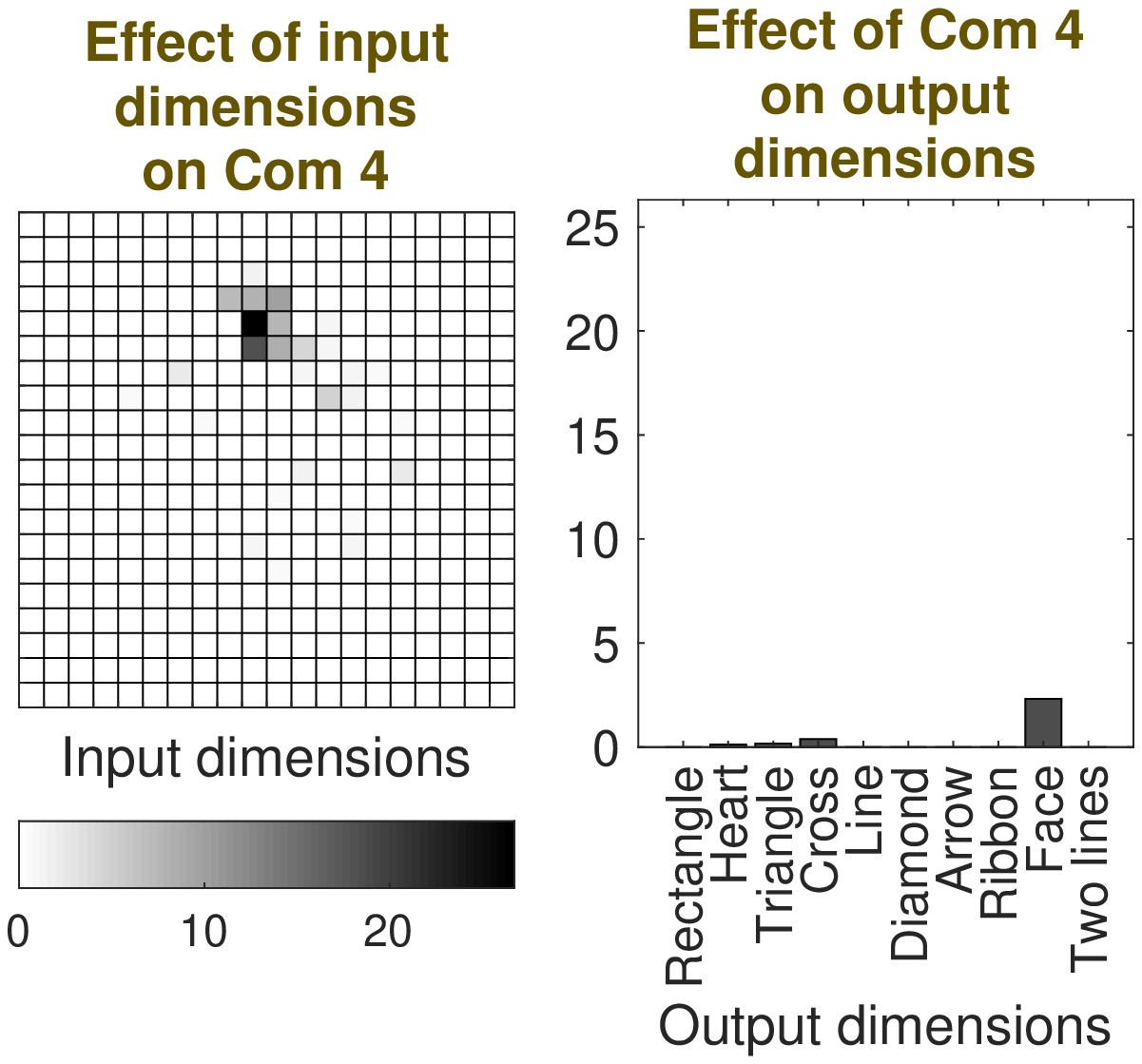}}
  \fbox{\includegraphics[height=33mm]{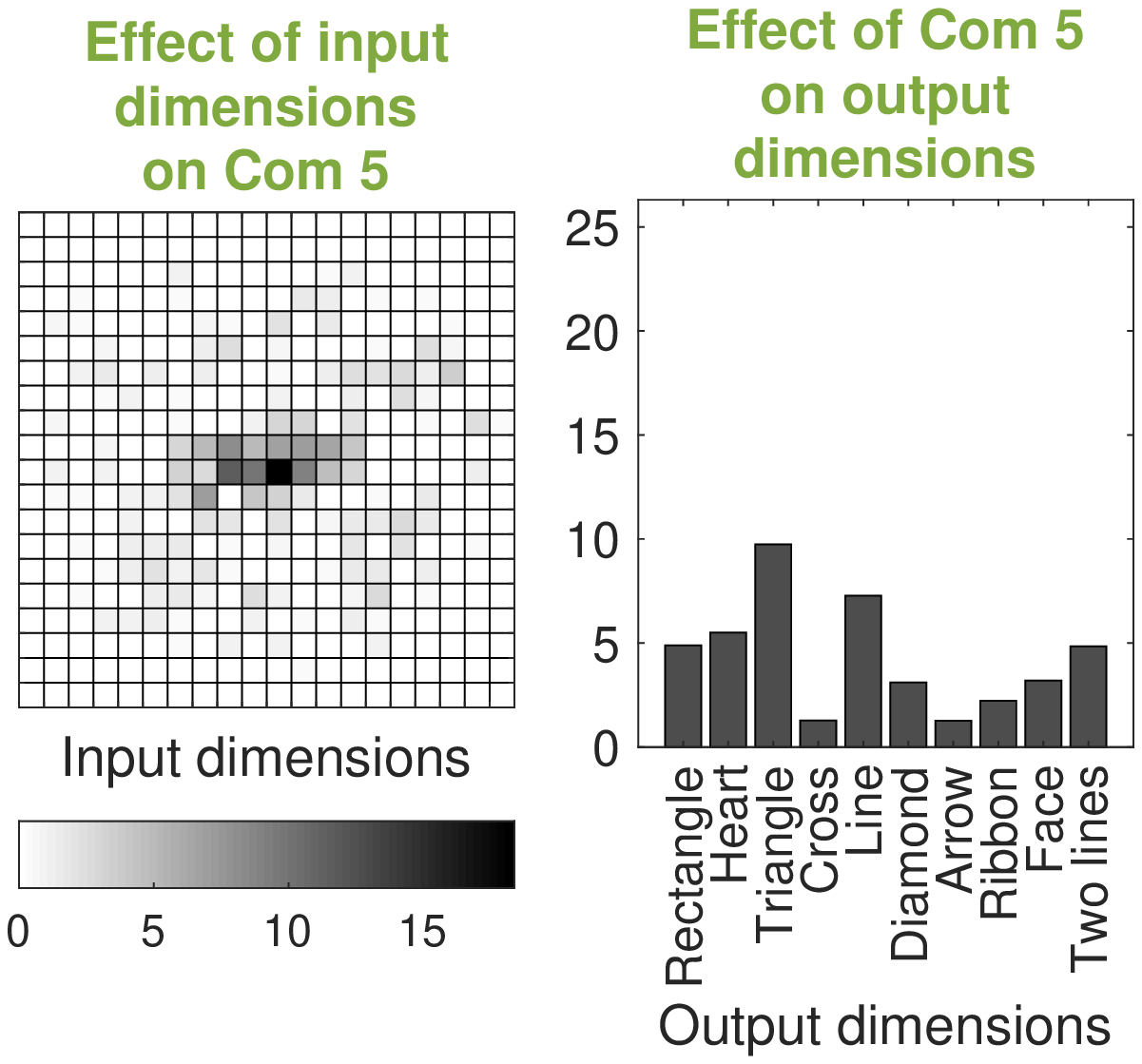}}
  \fbox{\includegraphics[height=33mm]{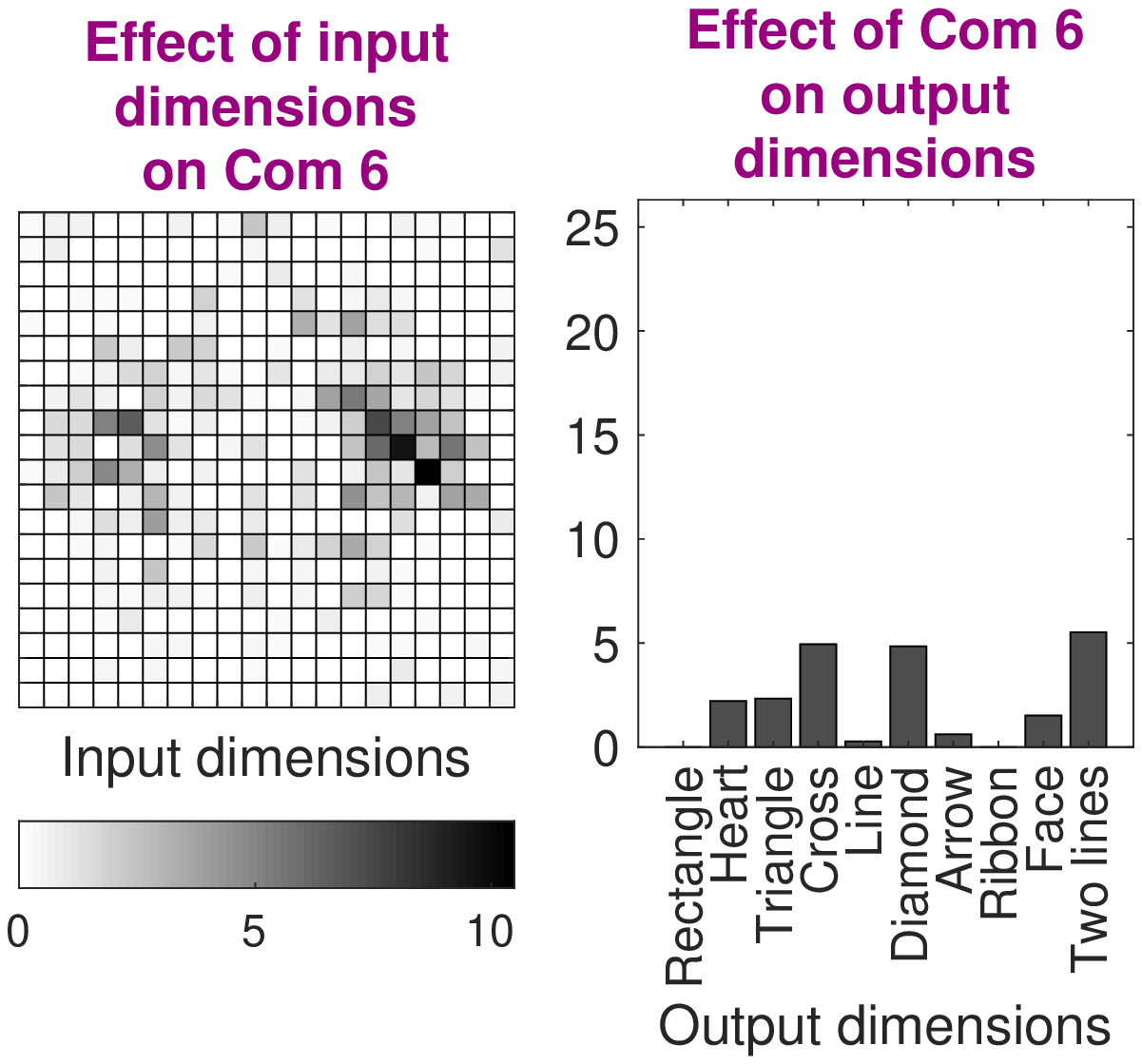}}\\
  %\hspace{10mm}
  \fbox{\includegraphics[height=33mm]{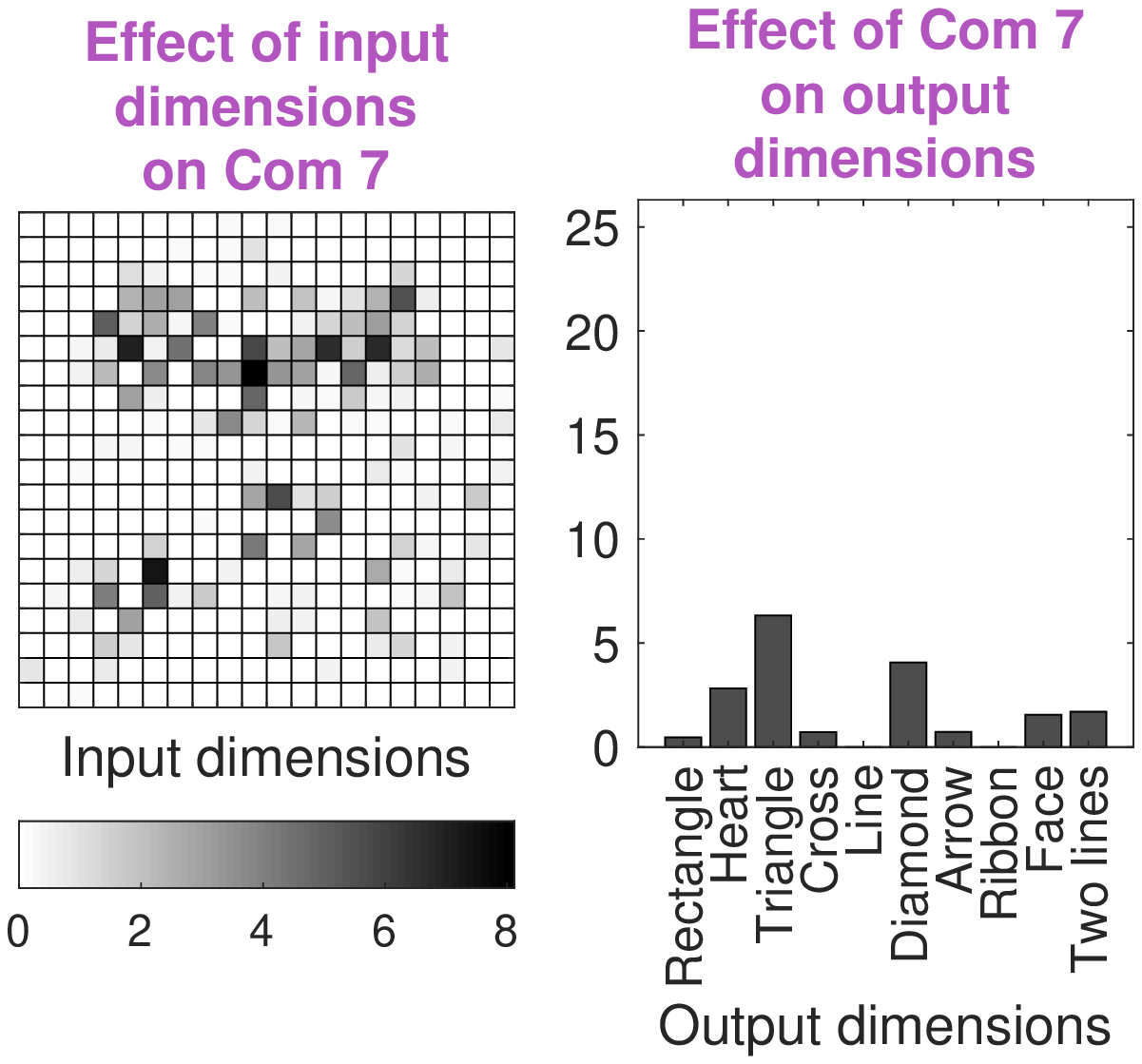}}
  \fbox{\includegraphics[height=33mm]{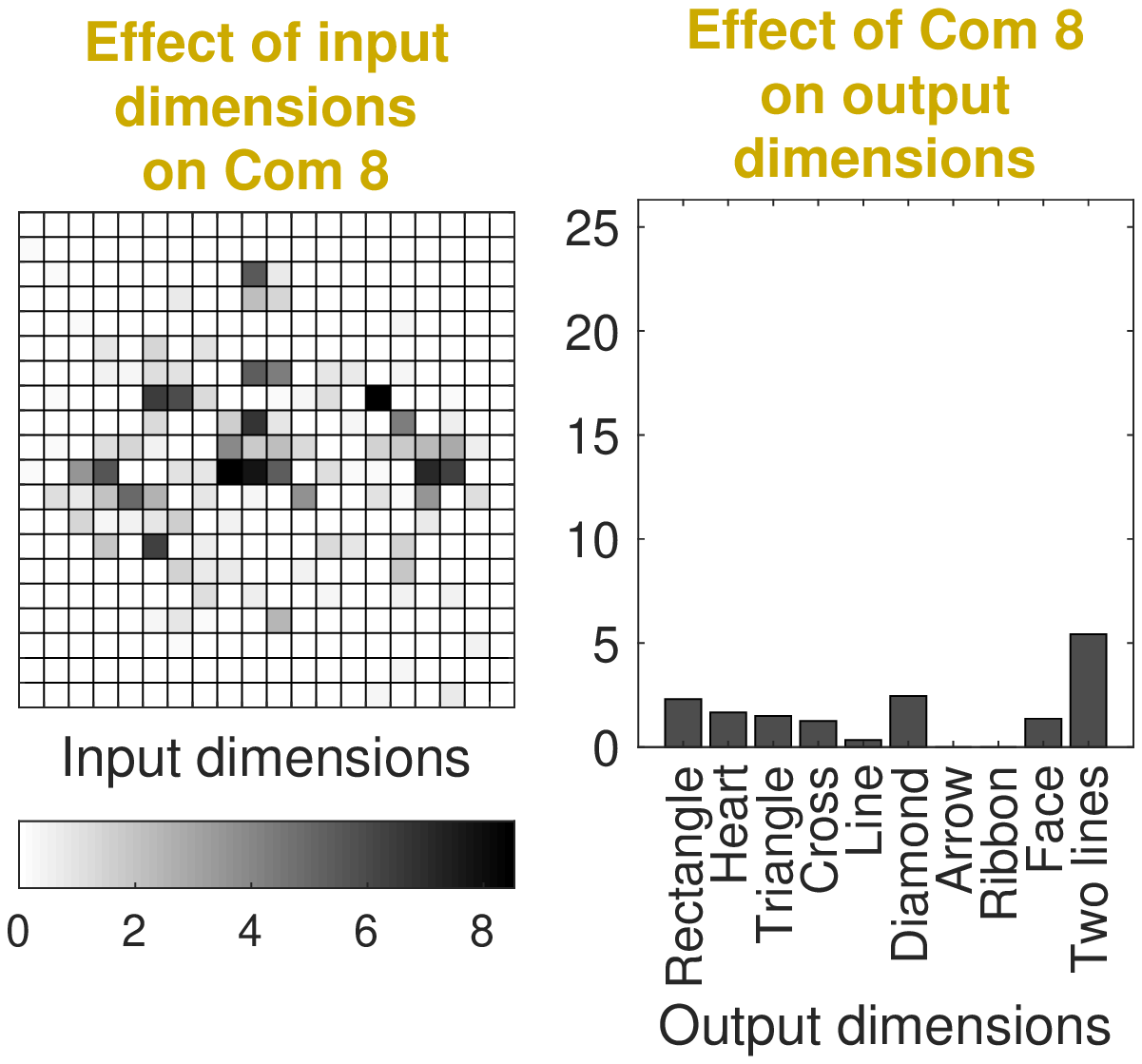}}
  \fbox{\includegraphics[height=33mm]{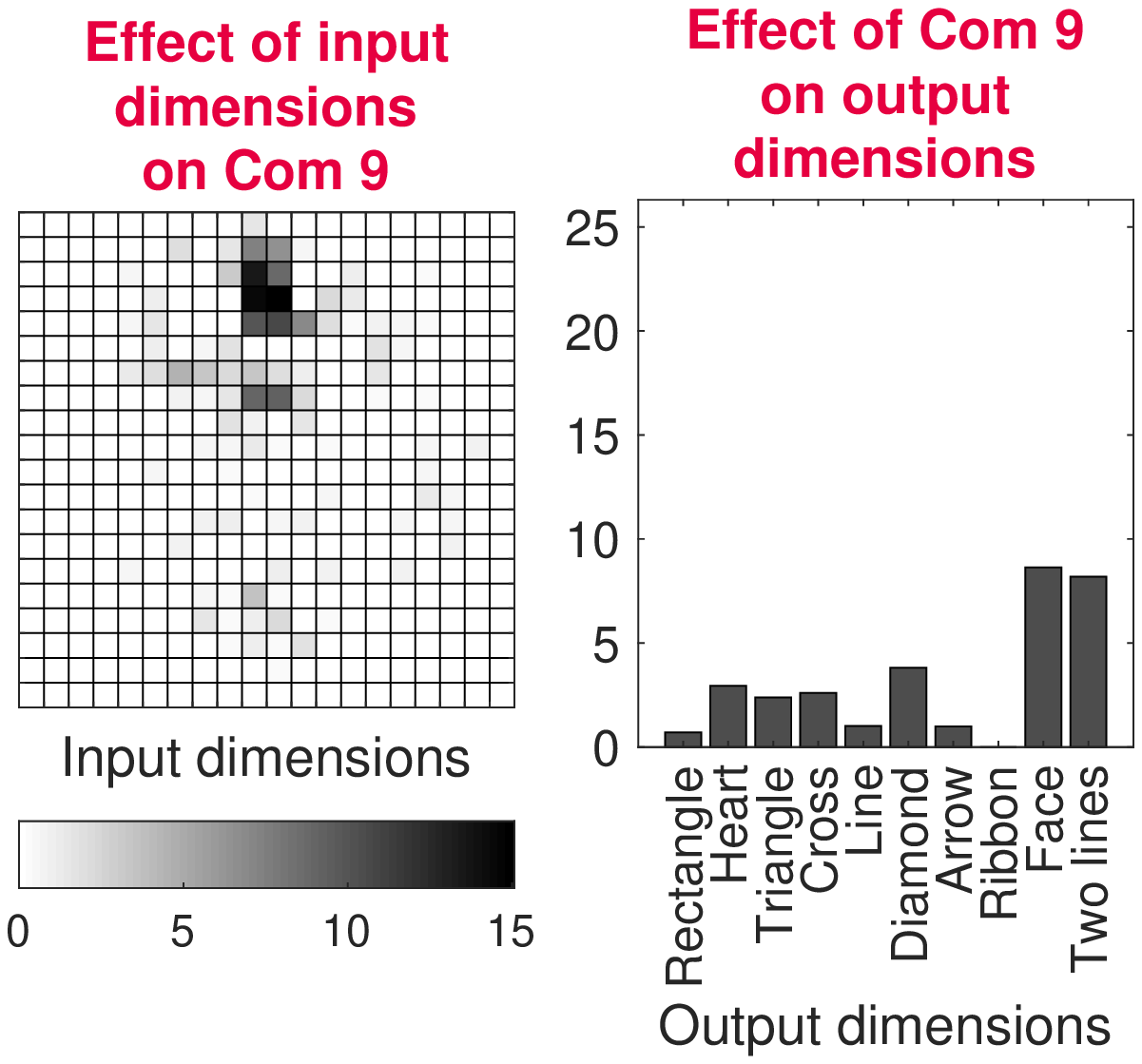}}\\
  %\hspace{10mm}
  \fbox{\includegraphics[height=33mm]{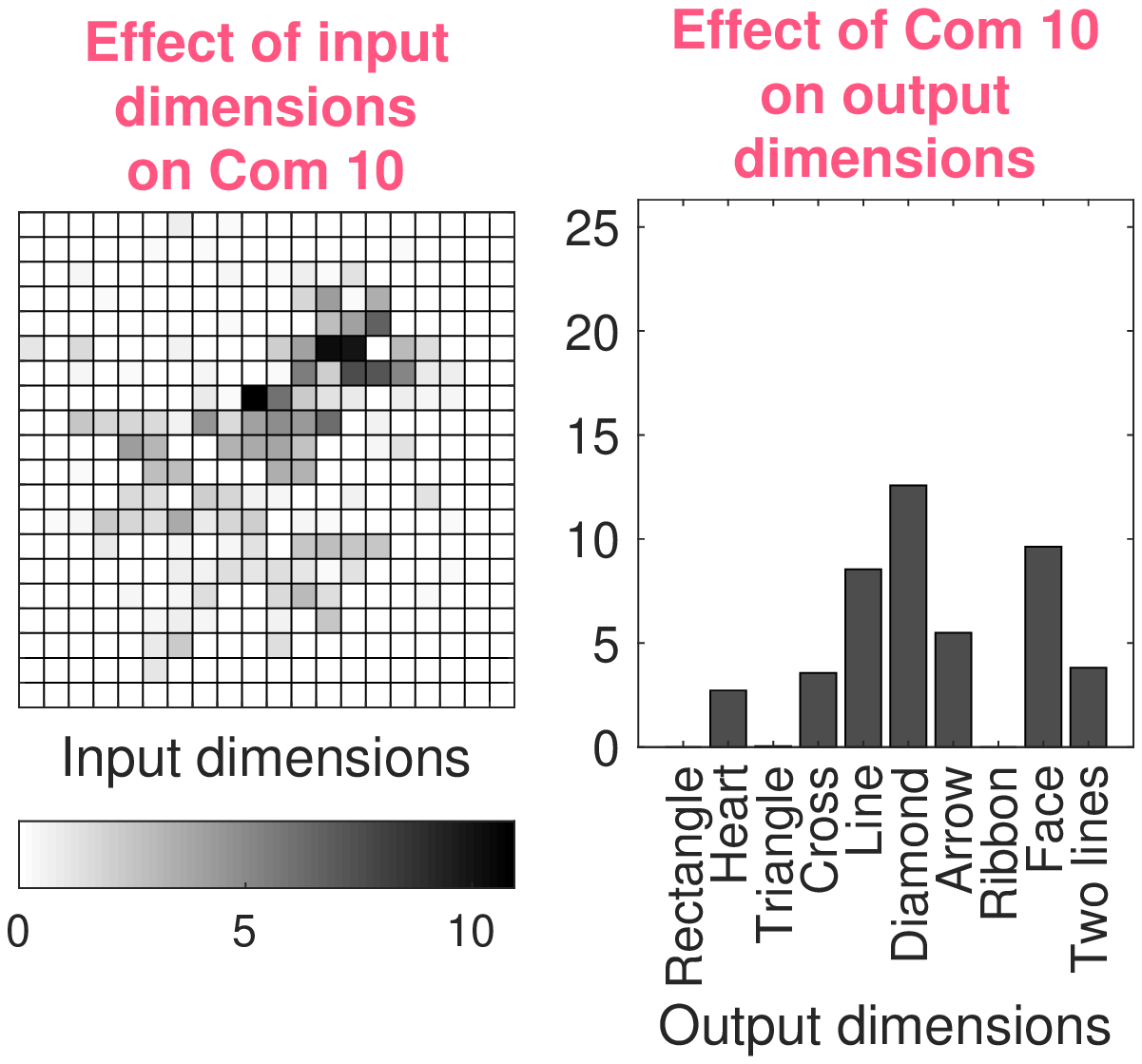}}
  \hspace{5mm}
  \includegraphics[height=33mm]{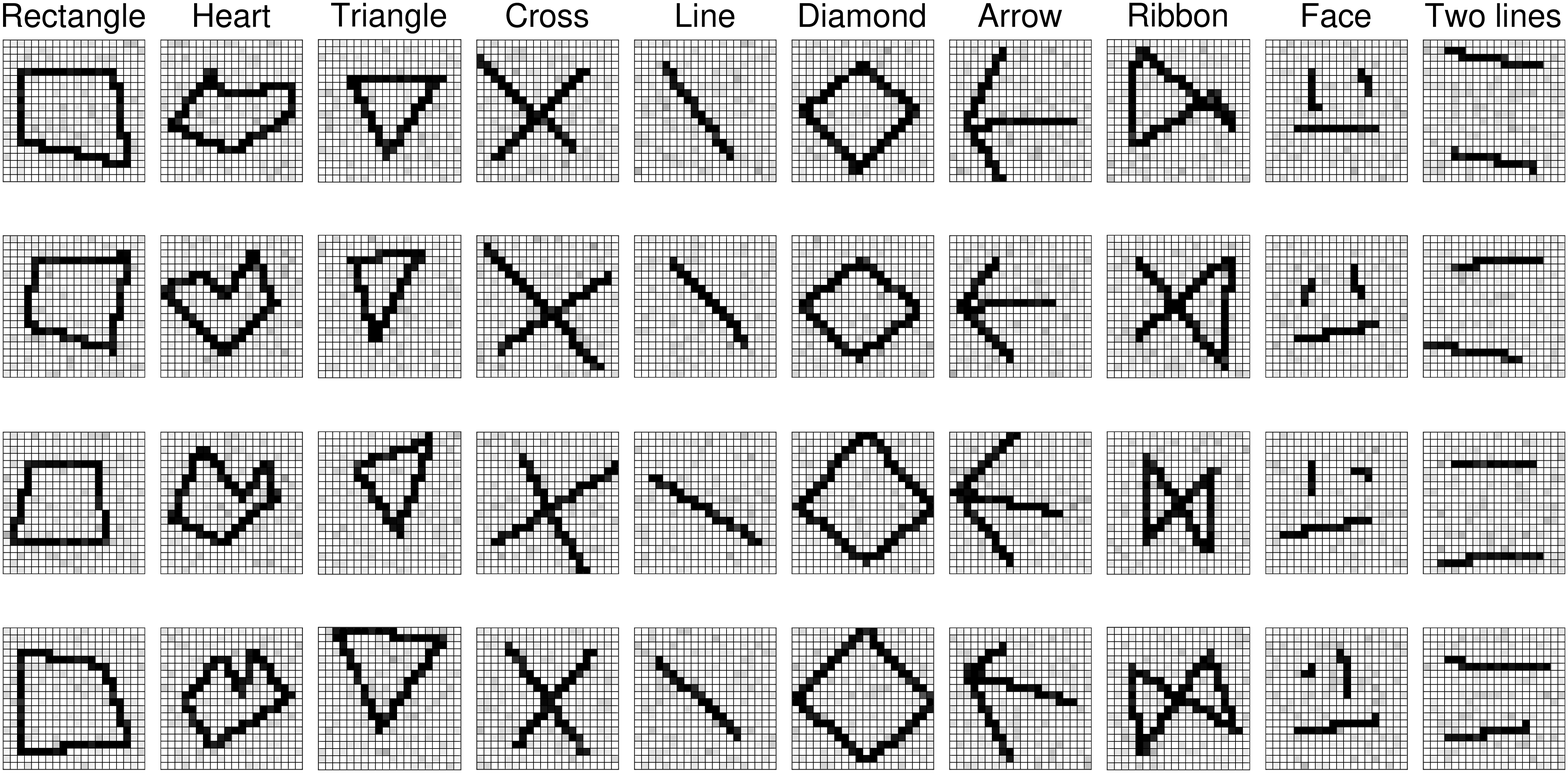}
  \caption{Decomposed tasks of an LNN trained with a diagram image data set. 
  The font colors of the figure titles correspond to the unit colors in Figure \ref{fig:diagram_lnn} (best viewed in color). Bottom right: Sample input image data for each class. }
  \label{fig:diagram_coms}
\end{figure}

Third, we applied our proposed method to an LNN trained with an image data set consisting of $20\times 20$ pixel images of $10$ types of diagrams 
that was also used in \cite{Watanabe2018arxiv}. The bottom right of Figure \ref{fig:diagram_coms} shows sample images for each class of diagrams. 
With this data set, we trained the LNN to recognize the $10$ types of diagrams from input images, and applied our proposed method to the trained LNN. 

Figures \ref{fig:diagram_lnn} and \ref{fig:diagram_coms}, respectively, show the trained LNN, and the decomposed tasks of the LNN. 
For instance, we can gain knowledge about the role of each community from these figures as follows. 

- Com $2$ is used for recognizing the ``Rectangle,'' ``Heart,'' and ``Diamond'' diagrams from the pixel information over a wide region, 
especially pixels near the left and right edges in the middle in a vertical direction, and pixels located in the upper left and right. 

- Com $3$ is used for recognizing the ``Cross'' and ``Ribbon'' diagrams mainly from the pixel information near the center of an image. 

- Com $5$ also uses the information given by pixels that are more localized in the center of an image than the pixels used by Com $2$, and it is used for recognizing various types of 
diagrams, such as ``Rectangle,'' ``Heart,'' ``Triangle,'' ``Line,'' and ``Two lines.'' 

- Com $9$ is used for recognizing ``Face'' and ``Two lines'' from the information provided by pixels located in the upper part of an image. 

- Com $10$ mainly recognizes the ``Line,'' ``Diamond,'' ``Arrow,'' and ``Face'' diagrams from the information given by an area extending from the upper right to the lower left 
in an image. 

%##################################################################################################################################################################################

\section{Discussions}

In this section, we discuss our proposed method in terms of both methodology and application. 
First, our proposed method enabled us to decompose an LNN into a given number of tasks, however, there is no statistical method for determining the number of tasks or 
communities. The construction of a criterion for optimizing the number of tasks is important future work. 

Next, in our proposed method, decomposed tasks are represented as non-negative vectors, and we cannot know what range of input dimension values results in what range of output 
dimension values. We need another method to reveal the role of each task in more detail. 

Finally, it would be possible to utilize our non-negative task decomposition to improve the generalization performance of an LNN. 
The experimental results in section \ref{sec:experiment} show that some communities have a bigger influence on the output dimensions than others. 
By observing such a disparity in the effect on prediction results, we can delete hidden units that are unimportant for inference and optimize the LNN architecture. 

%##################################################################################################################################################################################

\section{Conclusion}

LNNs have greatly improved predictive performance with various practical data sets, however, their inference mechanism has been black-boxed and 
we cannot interpret their complex training results. 
In this paper, we proposed a method for decomposing the function of hidden units in a trained LNN into multiple tasks, based on non-negative matrix factorization. 
We showed experimentally that our proposed method can provide us with knowledge about the role of each part of an LNN, in terms of which parts of the input and 
output dimensions are mainly related to them in relation to inference. 

%##################################################################################################################################################################################

%o o o o o o o o o o o o o o o o o o o o o o o o o o o o o o o o o o o o o o o o
%o o o o o o o o o o o o o o o o o o o o o o o o o o o o o o o o o o o o o o o o
%o o o o o o o o o o o o o o o o o o o o o o o o o o o o o o o o o o o o o o o o

%##################################################################################################################################################################################

\clearpage
\bibliographystyle{plain}

\end{document}